\documentclass[letterpaper, 10 pt, conference, nofonttune]{ieeeconf}
\IEEEoverridecommandlockouts
\usepackage{microtype}

\usepackage{cite}
\usepackage[bookmarks=true]{hyperref}
\usepackage{color}
\usepackage{times}
\usepackage{epsfig}
\usepackage{epstopdf}
\usepackage{graphicx}
\usepackage{amsmath}
\usepackage{amssymb}
\usepackage{bbm}
\usepackage[ruled,vlined,linesnumbered]{algorithm2e}
\usepackage{graphicx}
\usepackage{verbatim}
\usepackage{booktabs}
\usepackage{multirow}
\usepackage{makecell}
\usepackage{subfigure}
\usepackage{colortbl,booktabs}
\usepackage{lscape}
\usepackage{url}
\usepackage{makecell}
\usepackage{xcolor,colortbl}

\newcommand{\R}{\mathbb{R}}
\newcommand{\One}{\mathbf{1}}

\newcommand{\III}{\mathbf{I}}
\newcommand{\onemu}{\frac{1}{\mu}}
\newcommand{\mutwo}{\frac{\mu}{2}}

\newcommand{\WWW}{\mathbf{W}}

\newcommand{\pp}{\mathbf{p}}
\newcommand{\cc}{\mathbf{c}}

\newcommand{\ww}{\mathbf{w}}

\newcommand{\SSS}{\mathbf{S}}

\newcommand{\DDD}{\mathbf{D}}

\newcommand{\XXX}{\mathbf{X}}

\newcommand{\vneg}{\vspace{-7pt}}

\title{\LARGE \bf Adaptation to Team Composition Changes for Heterogeneous \\ Multi-Robot Sensor Coverage}

\author{
    Brian Reily, Terran Mott, and Hao Zhang%
    \thanks{*This work was partially supported by NSF CAREER Award IIS-1942056.}%
    \thanks{Brian Reily, Terran Mott, and Hao Zhang are with the Human-Centered Robotics Laboratory in the Department of Computer Science
    at the Colorado School of Mines, Golden, CO, 80401, USA.
    email: \{breily, terranmott, hzhang\}@mines.edu}%
}

\begin{document}

\maketitle
\thispagestyle{empty}
\pagestyle{empty}

\begin{abstract}

We consider the problem of multi-robot sensor coverage, which deals with deploying a multi-robot team in an environment and optimizing the sensing quality of the overall environment. As real-world environments involve a variety of sensory information, and individual robots are limited in their available number of sensors, successful multi-robot sensor coverage requires the deployment of robots in such a way that each individual team member's sensing quality is maximized. Additionally, because individual robots have varying complements of sensors and both robots and sensors can fail, robots must be able to adapt and adjust how they value each sensing capability in order to obtain the most complete view of the environment, even through changes in team composition. We introduce a novel formulation for sensor coverage by multi-robot teams with heterogeneous sensing capabilities that maximizes each robot's sensing quality, balancing the varying sensing capabilities of individual robots based on the overall team composition. We propose a solution based on regularized optimization that uses sparsity-inducing terms to ensure a robot team focuses on all possible event types, and which we show is proven to converge to the optimal solution. Through extensive simulation, we show that our approach is able to effectively deploy a multi-robot team to maximize the sensing quality of an environment, responding to failures in the multi-robot team more robustly than non-adaptive approaches.

\end{abstract}

\section{Introduction}

Multi-robot sensor coverage is the problem of deploying a team
of robots in an environment in order to maximize the observation
of events or phenomena \cite{cortes2002coverage,schwager2006distributed}.
Distributing coverage of an environment among the members of a 
multi-robot team allows their heterogeneous capabilities to 
be fully realized, with robots capable of sensing specific
events moving to the best positions possible for that particular
sensing modality.
In order to maximize the overall capability of a multi-robot 
team, they must be enabled to deploy themselves in such a way
that balances their individual sensing capabilities.
Effectively solving this problem is crucial for the deployment
of multi-robot systems to address real-world applications
such as search and rescue \cite{zadorozhny2013information} and 
security and surveillance \cite{meguerdichian2001exposure}.

\begin{figure}[t]
    \centering
    \vspace{6pt}
    \includegraphics[width=0.48\textwidth]{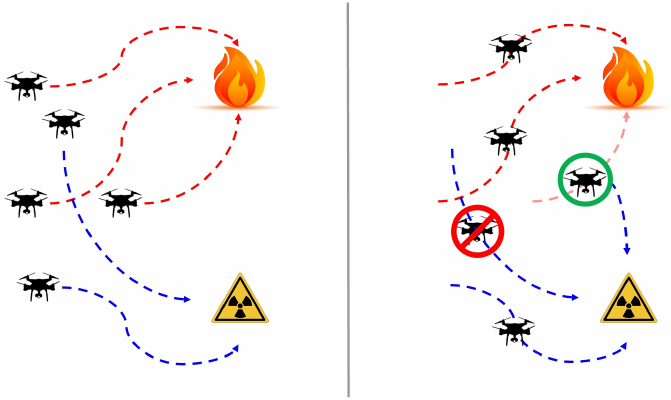}
    \vneg
    \caption{
    A motivating example of adaptation to team composition changes
    for heterogeneous multi-robot 
    sensor coverage. 
    In heterogeneous multi-robot sensor coverage, multiple robots
    are tasked with sensing multiple types of events.
    On the left, three robots begin moving towards the fire to
    improve their observations of it, while two robots move towards
    the radiation source.
    The right side shows the robots moving towards the events,
    and a robot has failed during operation.
    Our adaptive approach enables other robot team members to change
    how they value their sensing capabilities, and one robot,
    highlighted in green,
    shifts its weights to adapt to the change in the overall team
    sensing capability and provide observations of the radiation
    in place of the failed robot.
    }
    \label{fig:motivation}
\end{figure}

As real-world environments consist of multiple types of events
and robots can possess only a limited variety of sensing capabilities,
individual robots must be capable of dynamically balancing
their sensor inputs in order to construct the most complete
view of the environment.
For example, in a disaster, a robot may possess the ability to
sense both fire and radiation, while a teammate possesses only
a radiation sensor.
A more complete view of this disaster environment would be gained
if the first robot focuses its attention on fire, while the 
second robot focuses on radiation.
As real-world environments can be chaotic, these overall team
capabilities can change, as sensors or entire robots fail.
Multi-robot teams must be able to adapt to changes that occur,
and continuously balance their available capabilities in a way that
provides the best overall observation of the environment.
Figure \ref{fig:motivation} shows a motivating example of this,
where a team of five robots is tasked with sensing and monitoring
fire and radiation in an environment.
As the robots operate, a robot that has been observing the 
radiation source fails.
An adaptive approach enables the multi-robot team to react to
this change in team capability, and a robot that had focused on
the fire now shifts towards sensing the radiation.

Because of its relevance to many real-world applications, multi-robot
sensor coverage has seen significant recent research.
Many early approaches focused on sensor coverage of environments
with only a single form of sensory information by
homogeneous robot teams \cite{cortes2004coverage, pierson2015adapting}.
However, this limited view of the problem fails to properly
address real-world environments with a multitude of event types.
Several methods have been proposed to address heterogeneous environments
where multiple types of events can occur and teams consist of
robots with mixtures of sensing capabilities,
basing coverage on mixtures of probability 
distributions \cite{luo2018adaptive},
information maximization \cite{sadeghi2019coverage},
and defined control laws \cite{santos2018coverage_b}.
These methods all have the drawback of determining their balance
of sensing capabilities through fixed parameters, as opposed
to balancing the sensing capabilities based on the environment
and the team composition (i.e., with possible changes
due to robot failure).

In this paper, we introduce a novel formulation of multi-robot sensor
coverage that integrates the competing utilities provided by 
multiple sensing capabilities into a unified framework.
We consider a heterogeneous team of robots, where each individual robot
possesses only a subset of possible sensing capabilities,
operating  in an environment where multiple types of events 
have occurred.
The team is tasked with maximizing the overall sensing quality of
these events.
We propose an approach based on regularized optimization that finds
weights to optimally balance the utilities corresponding
to the various sensing capabilities, with an iterative
solver proven to converge to the optimal solution.
At each point in time, our approach identifies the optimal action
for each member of the multi-robot team, allowing each individual
robot to balance its competing sensing capabilities in order
to maximize its overall sensing quality and enabling the team
as a whole to adapt to changes in the environment and team composition.

We introduce two important contributions:
\begin{itemize}
\item We propose a novel formulation of heterogeneous
multi-robot sensor coverage, integrating multiple sensing capabilities
into a unified mathematical framework
based on regularized optimization.
Our formulation identifies an optimal balance between these
competing utilities, and does this at each time step, enabling
adaptation to changes in the environment and the composition
of the robot team.
\item We introduce an iterative algorithm to solve this proposed
problem, which is hard to solve due to non-smooth terms.
We show that this algorithm is theoretically proven to converge to
the optimal solution.
\end{itemize}

\section{Related Work} \label{sec:related}

As multi-robot sensor coverage has connections to many real-world
robotics applications, it is an active research area with multiple
approaches that address various aspects of it.
The key divisions of research are sensor coverage approaches in
homogeneous systems and heterogeneous systems.

Homogeneous sensor coverage addresses environments with only a
single type of event or a multi-robot system with the
capability to only sense a single event modality.
Accordingly, most homogeneous sensor coverage approaches address
the problem of evenly distributing multiple robots spatially
in an environment,
as there is no need to consider individual 
capabilities \cite{rekleitis2004limited,batalin2002spreading}.
This has been accomplished through Voronoi 
distributions \cite{guruprasad2013performance},
decomposition of an environment into cells \cite{hazon2005redundancy},
estimating density functions \cite{lee2015multirobot},
or representing an
environment as a graph and utilizing graph partitioning methods
to assign robots to regions
\cite{yun2012distributed,yun2014distributed,kong2006distributed}
or teams \cite{reily2020representing}.
Additionally, partitioning an environment has been done by
calculating the information gain estimated from different
regions \cite{fung2019coordinating},
using a market-based system to assign robots
based on information gain \cite{zlot2002multi},
or by planning paths that
use greedy algorithms to maximize spatial coverage
\cite{corah2017efficient}.

Homogeneous sensor coverage has also been extensively 
studied with the addition
of real-world constraints.
Maintaining communication is important for the success of multi-robot
operations, and multiple methods have focused on the deployment
of robots with constraints on communication
\cite{amigoni2019online, banfi2016asynchronous, banfi2017multirobot, penumarthi2017multirobot, reily2020leading}.
These methods have been based on both line of sight and distance
thresholds, and have been applied to open environments and obstructed
ones such as hallways.
Methods have also examined the physical limitations of sensors
and attempted to incorporate this into the control laws that dictate
their coverage approaches.
For example, visibility constraints of cameras
\cite{kantaros2015distributed},
limited range sensors \cite{pimenta2008sensing},
or limited field of view sensors \cite{gusrialdi2008coverage}
have all been integrated into deployment methods.
These constraints provide realistic representations of events and
sensing.
Finally, physical limitations on the robots themselves have also been
studied.
Power limitations were studied in \cite{kwok2010deployment,wang2012coverage},
where the real-world limitations on mobile robot batteries were
used to constrain the area that a team could cover.
Limitations on motors \cite{pierson2015adapting} and 
effects of traction and slippage \cite{pierson2017adapting}
also have been used to analyze paths to
coverage positions.
Turning radius was used as a key constraint in multiple works,
particularly with maneuverability-restricted robots such as
boats
\cite{notomista2019sensor,vandermeulen2019turn,karapetyan2018multi}.

While these various approaches towards homogeneous multi-robot
sensor coverage have been effective, they have the key limitation
of addressing only a single sensing modality, which is a poor
representation of real-world uses and applications.
To address this, heterogeneous multi-robot sensor coverage attempts
to solve the problem of coverage of multiple event types with a
multi-robot team that possess multiple forms of sensors.
Small multi-robot systems have been enabled to do this through
fixed scheduling algorithms \cite{shkurti2012multi},
following (e.g., robots
with heterogeneous capabilities move together so each can 
provide a perspective based on their sensor complement)
\cite{hood2017bird},
or integrating observations from robots performing 
other tasks \cite{gao2020regularized}.

For larger multi-robot teams, most approaches have utilized different
methods to optimize the `combined sensing quality', or the 
total information available to sensors across the possible event
types \cite{santos2018coverage_b}.
This has been done by optimizing a cost function
\cite{santos2018coverage}
or by identifying a distribution of robots that matches an
estimated sensing quality function \cite{sadeghi2019coverage}.
Voronoi regions have also been applied here, with their boundaries
based on multiple event types as opposed to a single one
\cite{arslan2016voronoi,guruprasad2013performance}.
These approaches are generally fixed, assuming a static event is
occurring and modeling robots as valuing each of their available
sensors equally.

Limited approaches have been proposed to adapt to dynamic changes
in the environment.
In \cite{luo2018adaptive},
robots learn a model of the events occurring from
sensor observations, and base their behavior on this model.
This has also been accomplished with a mixture of density functions
to model complex events \cite{schwager2009decentralized},
or by making online 
estimations of information gain in various parts of the 
environment \cite{schwager2006distributed}.
However, even these methods lack the ability to adapt to changes
in the team capabilities, and so are unable to respond to sensor
or robot failures.

In contrast to these reviewed approaches, our novel approach
to heterogeneous multi-robot sensor coverage is able to balance
available sensing capabilities in order to provide a more complete 
view of the environment.
Additionally, our formulation allows a multi-robot team to adapt
to changes in the environment and team composition, responding
to sensor or robot failures.

\section{Our Proposed Approach} \label{sec:approach}

In this section, we introduce our novel approach to heterogeneous
multi-robot sensor coverage that balances sensing quality based
on the capabilities available to each robot.  
We denote matrices with uppercase bold 
letters and vectors as lowercase bold letters.
Given a matrix $\XXX = [ x_{ij} ] \in \R^{n \times m}$, we denote
its $i$-th column as $\mathbf{x}_i$ and its $j$-th row as
$\mathbf{x}^j$.

\subsection{Problem Formulation}

We address the problem of a heterogeneous multi-robot team
tasked with covering an environment where multiple event types occur.
We define $N$ robot team members, each located
at a position denoted as $\pp_i$
for the $i$-th robot.
Each robot has a set of sensing capabilities 
denoted by a vector $\cc_i \in \R^E$, where
$E$ is the number of possible event types and $c_{ij} = 1$ if the
$i$-th robot has the $j$-th sensing capability, and $0$
otherwise.
We additionally define events occurring in the environment,
modelling each event with one or more density functions centered on
one or more positions (e.g., in a disaster scenario, smoke may be
spreading from a single fire or from several).
Events can be any of $E$ different types, corresponding to the set of
available sensing capabilities.

Each robot estimates the density functions corresponding to the
events based on its own observations of them,
where $\phi_{i,j} ( \cdot )$ denotes the $i$-th robot's estimation
of the $j$-th event type and returns a scalar value for a given
position.
At each time step, each robot incorporates sensor observations
at its position based on its heterogeneous capabilities
and updates the corresponding $\phi ( \cdot )$ functions.
As a robot moves towards a source of an event, the value returned
by the associated $\phi ( \cdot )$ rises; similarly, if a robot
were to move away from a source of an event the value would fall.
If a robot does not possess the necessary sensing capability
(i.e., $c_{ij} = 0$) then $\phi_{i,j} = 0$ for all positions.

To quantify the value of each of a robot's heterogeneous sensing
capabilities, we define the utility associated with moving 
towards an event type,
and thus increasing the sensing quality with respect to it.
We introduce $\SSS = [ s_{ij} ] \in \R^{N \times E}$
as the utility associated with sensing quality, where $s_{ij}$
describes the value of the $i$-th team member moving in the direction
of the $j$-th event type, given its current estimate of that event.
This utility is calculated using the gradient of $\phi_{i,j}$ 
with respect to the robot's current position $\pp_i$.
We denote the movement implied by this gradient as 
$\pp_{i \to j}$, which is a movement in the direction of the 
$j$-th event, based on the $i$-th robot's estimate of that event.
Formally, the utility is based on the value returned by
$\phi_{i,j}$ if this movement is taken:
\begin{equation}
\SSS = [ s_{ij} ] = %
\begin{cases}
\phi_{i,j} ( \pp_i + \pp_{i \to j} ) & \text{if } c_{ij} = 1 \\
0 & \text{if } c_{ij} = 0
\end{cases}
\end{equation}
We note that just as the utility $s_{ij} = 0$ if the $i$-th robot
cannot sense the $j$-th event type, the movement $\pp_{i \to j}$
is also equal to the zero vector.

Given the described utility $\SSS$, the objective of our problem
formulation is to maximize the overall sensing utility
based on each robot's current estimation of the events occurring
in the environment.
Each robot, given the utility of its various capabilities, must
find an optimal balance among them.
We apply this balance to the possible actions for each robot,
generating movements that allow them to maximize their individual
sensing quality based on their available capabilities.

\subsection{Optimization to Balance Sensing Capabilities}

We introduce an optimization-based formulation to identify 
an optimal balance of the competing utilities of the various 
sensing capabilities.
First, we introduce the base objective function, where we
maximize the overall utility provided by $\SSS$:
\begin{align}
\max_{\WWW} \; \| \WWW \odot \SSS \|_1
\end{align}
where $\odot$ denotes element-wise matrix multiplication
and $\| \cdot \|_1$ denotes the element-wise $\ell_1$-norm
of a matrix.
We introduce $\WWW \in \R^{N \times E}$,
which weights the sensing utilities, with $w_{ij}$ specifically 
representing the weight that the $i$-th robot assigns
to sensing the $j$-th event type.

To control the formation of this weight matrix, we introduce
the following constraints:
\begin{align}
\WWW \One_E = \One_N & \qquad \WWW \geq 0
\end{align}
where $\One_N$ is a vector of $1$s of length $N$.
We introduce these constraints to ensure that all weights
are positive and so that the weights assigned to each
individual robot in $\WWW$ sum to $1$ (i.e., no weight for
a sensing capability can grow unreasonably large).

Next, we introduce a regularization term to encourage the 
assignment of at least one robot to each event type in the environment.
To do this, we introduce the $\ell_2$-norm on each column of
$\WWW$ and define the \emph{event} norm:
\begin{align}
\| \WWW \|_E = \sum_{e=1}^E \| \ww_e \|_2
\end{align}
Because of the constraints introduced above, the values in
$\WWW$ are bounded to be between 0 and 1, and each row sums
to 1.
Maximizing this norm encourages values to form in each column
of $\WWW$, meaning that each event receives weights from a robot.
Otherwise, multiple robots could assign the maximum weight of
1 to a single event, leaving others unattended.

We also introduce a regularization term to enforce temporal
consistency in the weight matrix.
We note that if a robot is moving in the direction of an
event in order to improve its sensing quality, abruptly
switching directions at the next time step
to move towards an alternative event
is not ideal; changes should be gradual so as to not lose
progress made towards improved sensing quality.
To enforce this, we introduce specifying $\WWW$ as
$\WWW_t$, indicating the weight matrix at time step $t$,
and add a penalty term based on the difference
between the value of $\WWW_t$
with the value at the previous time step, $\WWW_{t-1}$:
\begin{align}
\| \WWW_t - \WWW_{t-1} \|_F^2
\end{align}
where $\| \cdot \|_F^2$ denotes the squared Frobenius norm.
We initialize $\WWW_0$ to give equal weights to all available
sensing modalities, i.e. if the $i$-th robot is capable of 
sensing $x$ of the $E$ possible sensing modalities, then each
entry $w_{ij} = \frac{1}{x}$.

Our final objective function combines these introduced
terms into a unified regularized optimization problem
that identifies an optimal balance between the competing
utilities of the available sensing capabilities:
\begin{align}
\max_{\WWW_t} & \; \| \WWW_t \odot \SSS \|_1 + \gamma_1 \| \WWW_t \|_E - \gamma_2 \| \WWW_t - \WWW_{t-1} \|_F^2 \notag \\
\text{s.t.} & \; \WWW_t \One_E = \One_N, \WWW_t \geq 0. 
\label{eq:final_obj}
\end{align}
where $\gamma_1$ and $\gamma_2$ are hyperparameters controlling
the importance of the two introduced regularization terms.

Earlier, we introduced $\pp_{i \to j}$, or the movement of the
$i$-th robot in the direction of the $j$-th event.
The overall movement $\dot{\pp}_i$ for the $i$-th robot is
based on a combination of these movements,
weighted by the weight matrix $\WWW_t = [ w_{ij} ]$
computed for time step $t$ in the objective function:
\begin{align}
\dot{\pp}_i = \sum_{j=1}^E w_{ij} \pp_{i \to j}
\end{align}
This overall movement update $\dot{\pp}_i$ is scaled to unit length
and added to the previous position $\pp_i$ to arrive at the new position:
\begin{align}
\pp_i = \pp_i + \frac{\dot{\pp}_i}{\| \dot{\pp}_i \|_2}
\end{align}

\subsection{Optimization Algorithm}

Because of the non-smooth terms and equality constraints, 
Eq. (\ref{eq:final_obj}) is hard to solve.
We propose an iterative solution based on the Augmented
Lagrangian Multiplier (ALM) method, similar to 
\cite{rockafellar1974augmented},
in which we can transform constraints
into penalty terms in the objective formulation.

\begin{algorithm}
    \caption{The general ALM method to solve Eq. (\ref{eq:alm})}
    \label{alg:alm}
    Set $1 < \rho < 2$ and initialize $\mu > 0$ and $\Lambda$.
    
    \While{not converge}
    {
    Update $\XXX$ by solving $\min_{\XXX} \; f ( \XXX ) + \frac{\mu}{2} \| h ( \XXX) + \frac{1}{\mu}\Lambda \|_F^2$;
    
    Update $\Lambda$ by $\Lambda = \Lambda + \mu h( \XXX )$\
    
    Update $\mu$ by $\mu = \rho\mu$\;
    }

\end{algorithm}

We consider problems of the form 
\begin{align}
\min f( \XXX ) \; \; \text{s.t.} \; h( \XXX ) = 0
\label{eq:alm}
\end{align}
Constrained optimization problems in this form can be solved by
the general ALM method described in Algorithm \ref{alg:alm}.
The equality constraint of $h ( \XXX ) = 0$ is transformed into
the penalty term added to $f( \XXX )$ in Line 3.
This line and the updates to $\mu$ and $\Lambda$ are repeated
until the value of $\XXX$ converges.

Following this general form, we can rewrite our final objective
function in Eq. (\ref{eq:final_obj}) and move the constraint of
$\WWW_t \One_E = \One_N$ into the objective function as a penalty
term.
At the same time, we rewrite our objective as a minimization problem
as opposed to maximization:
\begin{align}
\min_{\WWW_t} & \; - \| \WWW_t \odot \SSS \|_1 - \gamma_1 \| \WWW_t \|_E + \gamma_2 \| \WWW_t - \WWW_{t-1} \|_F^2 \notag \\
& \; + \mutwo \| \WWW_t \One_E - \One_N + \onemu \lambda \|_F^2 \\
\text{s.t.} & \; \WWW_t \geq 0. \notag
\label{eq:final_obj_rewrite}
\end{align}
where $\mu$ and $\lambda$ are introduced as multiplier variables.

We also note that $\| \WWW_t \odot \SSS \|_1$,
the element-wise $\ell_1$-norm of the Hadamard product,
or element-wise matrix multiplication, can be rewritten as the
Frobenius inner product, which is equal to the trace of the 
matrix product.
For the first term in our objective function, this means that
\begin{align}
\| \WWW_t \odot \SSS \|_1 = \langle \WWW_t, \SSS \rangle_F = \text{trace} ( \WWW_t^\top \SSS )
\end{align}
This makes our actual objective function
\begin{align}
\min_{\WWW_t} & \; - \text{trace} ( \WWW_t^\top \SSS ) - \gamma_1 \| \WWW_t \|_E + \gamma_2 \| \WWW_t - \WWW_{t-1} \|_F^2 \notag \\
& \; + \mutwo \| \WWW_t \One_E - \One_N + \onemu \lambda \|_F^2 \\
\text{s.t.} & \; \WWW_t \geq 0. \notag
\end{align}

To solve this rewritten objective function, we take the derivative
with respect to $\WWW_t$ and set it equal to 0:
\begin{align}
& -\SSS - \gamma_1 \WWW_t \DDD + 2 \gamma_2 \WWW_t - 2 \gamma_2 \WWW_{t-1}  \notag \\
& + \mu \WWW_t \One_E \One_E^\top - \mu \One_N \One_E^\top + \lambda \One_E^\top = 0
\label{eq:deriv_equal_0}
\end{align}

Here, $\DDD$ is a diagonal matrix such that
\begin{align}
\DDD = [ d_{ii} ] = \frac{1}{2 \| \ww_i \|_2}
\end{align}

After rearranging Eq. (\ref{eq:deriv_equal_0}), we see that the update
to $\WWW_t$ at each step is:
\begin{align}
\WWW_t = & \Bigg( \SSS + 2 \gamma_2 \WWW_{t-1} + \mu \One_N \One_E^\top - \lambda \One_E^\top \Bigg) \notag \\
& \Bigg( -\gamma_1 \DDD + 2 \gamma_2 \III + \mu \One_E \One_E^\top \Bigg)^{-1}
\label{eq:update_W1}
\end{align}

Finally, to ensure the $\WWW_t \geq 0$ constraint is incorporated,
we threshold the values in $\WWW_t$:
\begin{align}
\WWW_t = \text{max} ( \WWW_t, 0 )
\label{eq:update_W2}
\end{align}

After updating $\WWW_t$, we also update $\mu$ and $\lambda$:
\begin{align}
& \mu = \rho \mu \\
& \lambda = \lambda + \mu ( \WWW_t \One_E - \One_N )
\end{align}
where $\rho$ is a value chosen such that $1 < \rho < 2$.
These steps are repeated until the value of $\WWW_t$ converges.
This process is formally defined in Algorithm \ref{alg:alg_ours}.

\begin{algorithm}[tb]
\small
\SetAlgoLined
\SetKwInOut{Input}{Input}
\SetKwInOut{Output}{Output}
\SetNlSty{textrm}{}{:}
\SetKwComment{tcc}{/*}{*/}
\BlankLine

Set $1 < \rho < 2$ and $k = 0$.
Initialize the penalty coefficient $\mu^0 > 0$ and the multiplier
term $\lambda^0$.
Initialize the weight matrix $\WWW_t$.

\Repeat{
convergence
}{
Compute $\DDD^k = \text{diag} \left( \frac{1}{2 \| \ww_i \|_2} \right)$.

Compute $\WWW_t^{k+1}$ by Eqs. (\ref{eq:update_W1}) and (\ref{eq:update_W2}).

Update $\lambda$ by $\lambda^{k+1} = \lambda^k + \mu^k ( \WWW_t^{k+1} \One_E - \One_N )$.

Update $\mu$ by $\mu^{k+1} = \rho \mu^k$.

$k = k + 1$.
}
\caption{Our Algorithm to Solve Eq. (\ref{eq:final_obj}).}
\label{alg:alg_ours}
\end{algorithm}

\textbf{Computational Complexity}.
In Algorithm \ref{alg:alg_ours}, Lines 3, 5, 6, and 7 are trivial
and can be computed in linear time.
The computational complexity of our proposed solution is determined
solely by Line 4, which computes both a matrix inverse and a matrix
multiplication.
Respectively, these have complexities of $\mathcal{O} ( E^3 )$
and $\mathcal{O} ( N E^2 )$.
Typically, $N$ will be much larger than $E$ (i.e., a scenario where
the number of possible event types exceeds the number of available
robots is not one that will be able to be comprehensively sensed,
and so the number of robots will need to increase).
When this is the case, the overall complexity of each iteration of
our proposed solution algorithm is
$\mathcal{O} ( N E^2 )$.

\textbf{Convergence}.
Under the condition that $0 < \mu^k < \mu^{k+1}$, the general ALM
approach described in Algorithm \ref{alg:alm} is proven to converge
to an optimal value of $\XXX$ \cite{bertsekas2014constrained}.
As we initialize $\mu^0 > 0$, then $0 < \mu^k$ holds at $k = 1$.
We also initialize the parameter $\rho$ such that $1 < \rho < 2$,
and this parameter controls the only update to $\mu$ in Line 6.
Thus, $\mu^{k+1}$ cannot be less than $\mu^k$, as this would require
that $\rho < 1$, and so $\mu^k < \mu^{k+1}$ holds at every step.


\section{Experiments} \label{sec:results}

\subsection{Experimental Setup}

In order to comprehensively evaluate our adaptive multi-robot sensor
coverage approach, we performed both extensive simulations
in a high-fidelity simulator to integrate real-world
control considerations.
This simulator also required our approach to integrate with the
Robot Operating System (ROS), as would be necessary on physical
robots.

We evaluate the effects on various combinations of multi-robot
team sizes ($N = \{ 5, 10 \}$) and numbers of event types
($E = \{ 2, 3, 4 \}$).
Evaluation is conducted with each event type being randomly generated at two positions.
Members of a multi-robot team are also initialized at 
randomly generated positions near a chosen
start area in this environment, with only
a subset of $E$ possible sensors available to each robot.
We conduct each simulation until $t = 75$.
In order to demonstrate the adaptive abilities of our approach,
we simulate various numbers of robot failures during the simulation.

In all evaluations, we considered the metric of \emph{sensing quality},
or the improvement of sensing performance over the base deployment
of the multi-robot system.
This metric relates the sensing quality at a specific point to 
the initial sensing quality when robots are randomly positioned
in an environment.
That is, if a system begins with robots deployed within an environment
and they each proceed intelligently based on their sensors, 
then the overall sensing quality would improve.
Specifically, we look at the improvement of sensing quality at the
end of the simulation.

As our approach is not only adaptive to the distribution of sensors
but the availability of sensors as a dynamic system proceeds (i.e.,
in the real world, robots can fail or lose sensing capabilities
due to environmental factors), we consider a number of alternate
approaches in order to demonstrate the effectiveness of our
approach.
We compare to three alternate approaches in order to evaluate 
our proposed method for multi-robot sensor coverage:

\begin{table*}
    \centering
    \caption{Sensing Quality Improvement}
    \label{tab:sensing quality}
    \vspace{-9pt}
    Results are reported as multiples of the initial sensing quality.
    E.g., an initial sensing quality of $0.50$ and a final sensing
    quality of $5.00$ will be reported as $10.00$.
    For each combination of $N$ and $E$, we report with results with
    0, 1, 2, and 3 robot failures.
    The best improvements are highlighted in \textbf{bold} text.
    \\
    \vspace{1pt}
    \tabcolsep=0.35cm
    \begin{tabular}{|c|c|c|c|c|}
    \hline
    \# of Robots ($N$) & Approach & 2 Events ($E = 2$) & 3 Events ($E = 3$) & 4 Events ($E = 4$) \\
    \hline \hline
    \multirow{4}{*}{$N = 5$} & Baseline & 9.99 / 6.69 / 5.75 / 2.88 & 8.85 / 8.05 / 5.12 / 3.24 & 10.02 / \textbf{9.00} / \textbf{5.50} / 2.23 \\
    & Equally Weighted & 8.87 / 5.69 / 4.94 / 2.77 & 9.47 / 7.61 / 4.99 / 2.89 & 6.47 / 5.50 / 3.73 / 2.38 \\
    & Single Capability & 9.66 / 6.68 / 6.18 / 2.71 & 7.88 / 6.45 / 4.96 / 2.83 & 6.37 / 3.96 / 2.98 / 1.60 \\
    & Our Full Approach & \textbf{12.94} / \textbf{11.44} / \textbf{7.22} / \textbf{4.30} & \textbf{11.55} / \textbf{11.45} / \textbf{6.26} / \textbf{5.15} & \textbf{10.84} / 8.41 / 5.27 / \textbf{3.92} \\
    \hline \hline
    \multirow{4}{*}{$N = 10$} & Baseline & 10.35 / 10.30 / 7.76 / 5.93 & \textbf{11.85} / 8.89 / 7.69 / 6.83 & 7.99 / 7.48 / 6.29 / 4.24 \\
    & Equally Weighted & 10.59 / 10.12 / 8.39 / 6.48 & 7.09 / 6.77 / 6.30 / 2.05 & 6.28 / 5.80 / 4.99 / 3.46 \\
    & Single Capability & 9.84 / 9.06 / 6.50 / 5.77 & 7.20 / 5.63 / 4.98 / 4.05 & 6.36 / 5.90 / 5.40 / 4.87 \\
    & Our Full Approach & \textbf{13.19} / \textbf{11.25} / \textbf{9.69} / \textbf{8.92} & 11.35 / \textbf{9.56} / \textbf{8.33} / \textbf{7.87} & \textbf{12.64} / \textbf{9.66} / \textbf{6.56} / \textbf{5.79} \\
    \hline
    \end{tabular}
\end{table*}


\begin{figure*}[h]
    \centering
    \subfigure[Initial State]{
        \label{fig:dcist_0}
        \centering
        \includegraphics[width=0.228\textwidth]{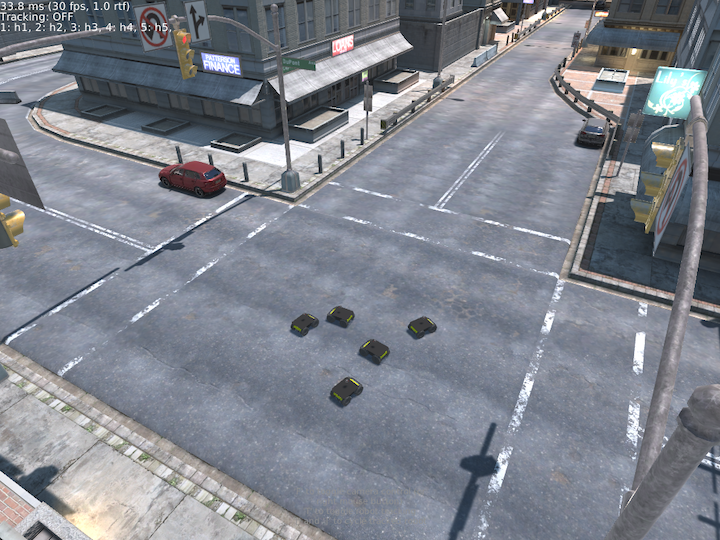}
    }%
    \subfigure[Deployment]{
        \label{fig:dcist_1}
        \centering
        \includegraphics[width=0.228\textwidth]{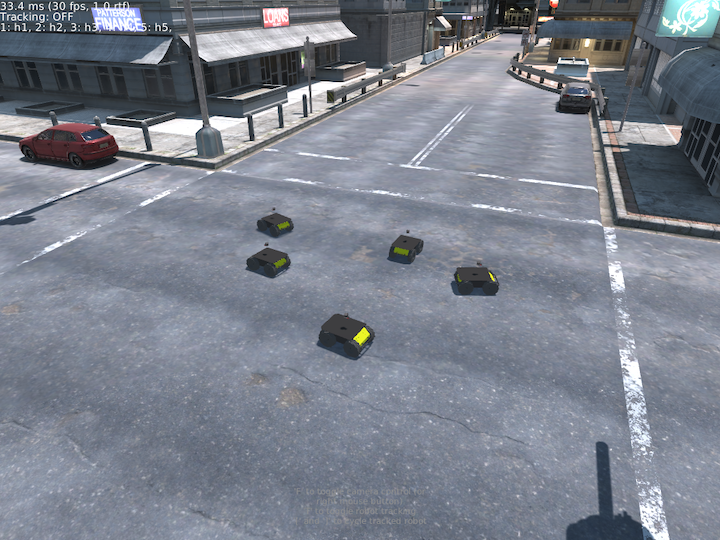}
    }%
    \subfigure[Failure]{
        \label{fig:dcist_2}
        \centering
        \includegraphics[width=0.228\textwidth]{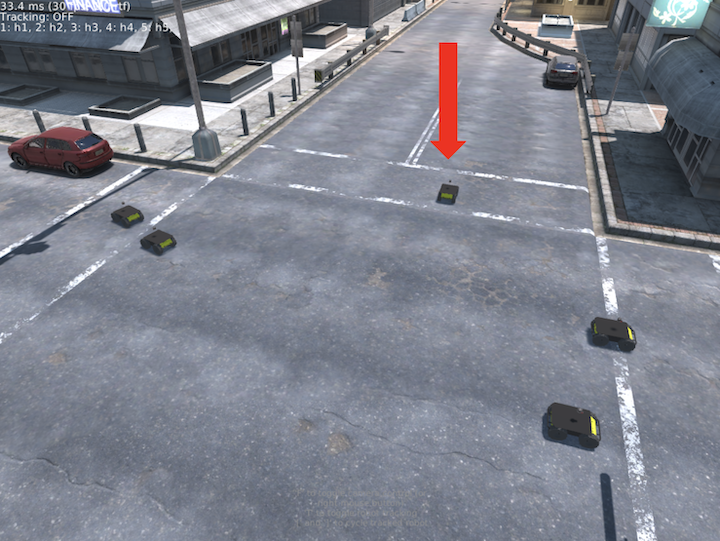}
    }%
    \subfigure[Adaptation]{
        \label{fig:dcist_3}
        \centering
        \includegraphics[width=0.228\textwidth]{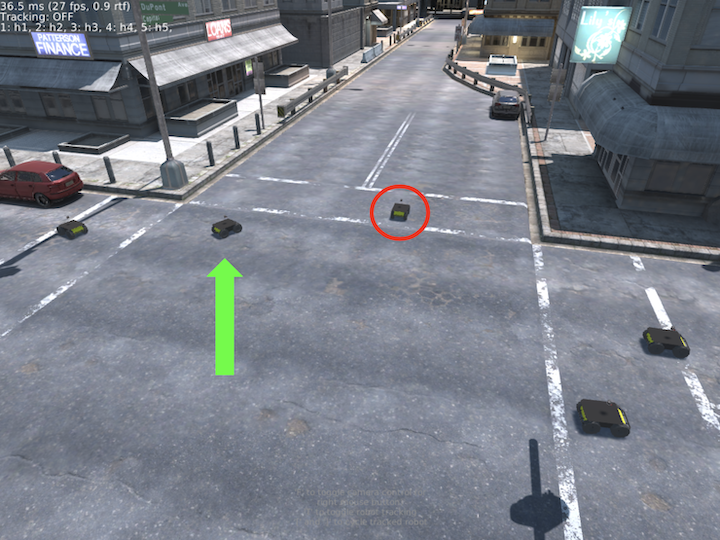}
    }
    \vneg
    \caption{These figures show a qualitative evaluation of our
    approach in a high-fidelity robot simulator.
    Figures \ref{fig:dcist_0} - \ref{fig:dcist_3} show the 
    progression of a multi-robot team as it moves through 
    a simulated high-fidelity environment.
    The robot team members individually travel to optimize their
    own sensing qualities, yet through our proposed approach they
    adapt to the available capabilities and react accordingly.
    When failure occurs in the robot marked with the red arrow
    in Figure \ref{fig:dcist_2}, another robot, marked with
    a green arrow in Figure \ref{fig:dcist_3} is able to adapt
    and shift towards sensing that event.
    }
    \label{fig:dcist}
\end{figure*}

\begin{enumerate}
    \item \emph{Baseline}: This approach sets $\gamma_1 = 0$ and
    $\gamma_2 = 0$, and so continues to find a weight matrix
    $\WWW$ that maximizes the available utility but does not
    utilize regularization to influence the development of $\WWW$.
    This approach still attempts to adapt each robot's weighting
    of its capabilities in order to provide a complete view of the
    environment.
    \item \emph{Equally Weighted}: This approach defines an equally
    weighted $\WWW$, where each robot assigns identical 
    values to each of its available sensing capabilities (i.e.,
    if a robot has two sensing capabilities, it
    assigns a weight of $0.5$ to each of them).
    This approach does not adapt to the availability of sensors
    or the failure of robots during the sensor coverage task.
    \item \emph{Single Capability}: This approach randomly 
    selects an available sensing capability for each robot and
    only allows the robot to use that sensor (i.e., if a 
    robot has an RGB camera and a depth camera, this
    approach limits the robot to only one and ignores the other).
    Similar to the \emph{Equally Weighted} approach, this approach does
    not adapt to changes in the system as the robots operate.
\end{enumerate}

\subsection{Evaluation on Simulated Multi-Robot Systems}

We present extensive quantitative results in 
Table \ref{tab:sensing quality}.
For each combination of $N = \{ 5, 10 \}$ and $E = \{ 2, 3, 4 \}$,
we conduct simulations with $0, 1, 2$, and $3$ robot failures.
Each combination of parameters and failures is simulated 100 
times.
We report the improvement in sensing quality at the end of
the simulation and at its highest point.
This is reported as a multiple of the initial sensing quality,
e.g. if the initial sensing quality is $0.50$ and the sensing quality
at the end of the simulation is $5.00$, we report an improvement
of $10.00$.

We observe that our full approach consistently provides the largest
sensing quality improvement, across nearly every combination of $N$, $E$, 
and number of robot failures.
This demonstrates the effectiveness of our approach to identify an 
optimal weighting of available sensing capabilities, assigning weights
in the context of the capabilities available to the overall team.
Additionally, we see that as the number of robot failures increases,
our approach widens its performance gap over the compared approaches,
indicating that its ability to adapt to changes in the multi-robot team
best enables it to overcome robot failure and continue to provide
effective sensing performance.
In some cases, our approach provides multiple times as much sensing
improvement as the compared approaches.
For example, for $N = 10$, $E = 3$, and three robot failures, our
approach is able to increase sensing quality 7.87 times the initial
value, while the \emph{Equally Weighted} approach only doubles it.
This shows the main strength of our approach, in that it is able
to adapt to changes in team composition (i.e., failure) and 
continue to optimally balance the remaining
available sensing capabilities.

In a few combinations, our baseline approach with $\gamma_1 = 0$ and
$\gamma_2 = 0$ slightly edges out our the full approach,
and when it does not it still 
consistently performs the second best or very near to it.
This shows that even without our introduced regularization terms that
distribute weights among event types and maintain temporal consistency,
our approach's ability to identify an optimal balance between sensing
capabilities is much more effective than relying on a either a single
capability or an equal weighting of capabilities.

Figure \ref{fig:dcist} shows qualitative results from
example simulation of multi-robot
sensor coverage in an urban environment.
The initial state is seen in Figure \ref{fig:dcist_0}, with five
Husky ground robots.
Three events are simulated, located down each road entering
the three-way intersection.
Figure \ref{fig:dcist_1} shows the ground robots deploying
towards the simulated events.
Two robots are moving towards the left road, one moving up
the road entering the top of the frame,
and the remaining two towards the road on the right.
Figure \ref{fig:dcist_2} shows the simulated failure, with the
Husky robot marked with the large red arrow failing.
As this was the only robot moving towards the event at the
top of the frame, existing approaches that cannot adapt would
lose observations of this event.
In Figure \ref{fig:dcist_3} we see that our approach is able
to adapt to this failure.
One of the robots that had been moving left has shifted its weighting
of its sensing capabilities and is now moving towards the top
of the frame to provide observations of that event.
Approaches that prioritize only a single sensing modality or
that do not adjust the weighting of sensing modalities would
not be able to adapt to this failure, leaving the event type
completely unobserved.

\section{Conclusion} \label{sec:conclusion}

Multi-robot sensor coverage is the problem of deploying a multi-robot
team in an environment in order to maximize the overall sensing quality.
Real-world environments consist of a variety of event modalities,
and so in order to provide a complete and comprehensive view of
an environment, a multi-robot team must deploy intelligently
based on its available sensing capabilities.
In addition, failures can occur to both sensors and robots, and
so a multi-robot team must be able to adapt to these, and change
its behavior to continue to provide high-quality sensing.
In this paper, we present a novel formulation of heterogeneous
multi-robot sensor coverage in which we provide an adaptive
approach based on regularized optimization.
We propose a problem formulation that integrates multiple
sensing capabilities and identifies an optimal balance of
these capabilities at each time step, adapting to not only
the available capabilities but also changes in the environment
and the multi-robot system.
We introduce an iterative algorithm to solve this formulated
problem, which we show is proven to converge to an optimal solution.
Through extensive simulation, we demonstrate that our approach 
provides effective multi-sensor robot coverage, outperforming
methods that focus on a single capability or that are unable to
adapt to changes in robot capabilities.

\clearpage
\bibliographystyle{ieeetr}
\bibliography{references}

\end{document}